\documentclass{article}

\PassOptionsToPackage{numbers, compress}{natbib}
 \usepackage[preprint]{neurips_2026}

\usepackage{graphicx}
\usepackage{enumitem}
\usepackage{wrapfig}
\usepackage{amsmath}
\usepackage{amsfonts}
\usepackage{multirow}
\usepackage[utf8]{inputenc} 
\usepackage[T1]{fontenc}    
\usepackage{hyperref}       
\usepackage{url}            
\usepackage{booktabs}       
\usepackage{amsfonts}       
\usepackage{nicefrac}       
\usepackage{microtype}      
\usepackage{xcolor}         
\usepackage{tcolorbox}
\newtcolorbox{prompt}[1]{
    left=4mm,
    right=4mm,
    top=2mm,
    bottom=2mm,
    boxsep=0mm,
    rounded corners,
    title=#1,
    fontupper=\footnotesize\linespread{0.95}\fontfamily{lmr}\selectfont,
    }
\title{Think-with-Rubrics: From External Evaluator to Internal         
  Reasoning Guidance}

\author{%
  Jiachen Yu\textsuperscript{1,}\thanks{Equal contribution.} \quad 
  Zhihao Xu\textsuperscript{2,}\footnotemark[1] \quad 
  Junjie Wang\textsuperscript{1,}\thanks{Correspondence to Junjie Wang <wangjunjie@sz.tsinghua.edu.cn> Yujiu Yang <yang.yujiu@sz.tsinghua.edu.cn>} \quad 
  Yujiu Yang\textsuperscript{1,}\footnotemark[2] \\
  \vspace{0.25cm} 
  \textsuperscript{1}Tsinghua University \qquad 
  \textsuperscript{2}Renmin University of China \\
}

\begin{document}

\maketitle

\begin{abstract}
  Rubrics have been extensively utilized for evaluating unverifiable, open-ended tasks, with recent research incorporating them into reward systems for reinforcement learning. However, existing frameworks typically treat rubrics only as external evaluator disjointed from the policy's primary reasoning trace. Such design confines rubrics to post-hoc measurement, leaving them unable to
  actively guide the model's generation process. In this work, we introduce \textbf{\textsc{Think-with-Rubrics}}, a novel paradigm for instruction following tasks. \textsc{Think-with-Rubrics} integrates rubric generation into the reasoning context, transforming the rubric from an independent artifact into an internal guidance of LLM's generation. During training, LLM sequentially generates a rubric followed by a response, while a trained rubric verifier provides joint supervision by evaluating the consistency between the answer and the self-generated / golden rubrics. Experiments across multiple benchmarks demonstrate that \textsc{Think-with-Rubrics} consistently outperforms the Rubric-as-Reward baseline supervised by golden rubrics by an average of 3.87 points. We have also discussed the mechanism by which \textsc{Think-with-Rubrics} enhances model performance. Experimental results demonstrate that supervision from golden rubrics and self-generated rubrics enhances the performance of \textsc{Think-with-Rubrics} by improving the quality of self-generated rubrics and increasing the internal consistency of responses respectively.
\end{abstract}

\section{Introduction}

Reinforcement learning (RL) has driven substantial progress in LLMs on verifiable tasks \citep{deepseek, deepseekmath}, where reliable reward signals can be derived from ground-truth labels.
However, extending RL to open-ended, unverifiable domains remains challenging, as such tasks lack ground truth.
In these settings, scalar rewards \citep{skywork,helpsteer} and monolithic generative reward models \citep{conj,rise} often provide supervision that is too coarse and underspecified to capture response quality.

To address this, recent research has increasingly focused on rubric-as-reward \citep{rar,openrubrics,sciif,checklistsbetterthan} paradigm: using structured sets of criteria to decompose each open-ended task into interpretable objectives. Rewards are then assigned based on how many rubric criteria a model’s response satisfies, providing more stable and informative supervision during optimization. Yet, in existing reward systems, these rubrics are only used as external evaluators after the model has finished its response. As a result, these success standards serve only to \textit{measure} the model's performance, not to \textit{guide} its generation.

\begin{figure}
  \centering
   \includegraphics[width=\linewidth,height=0.26\textheight]{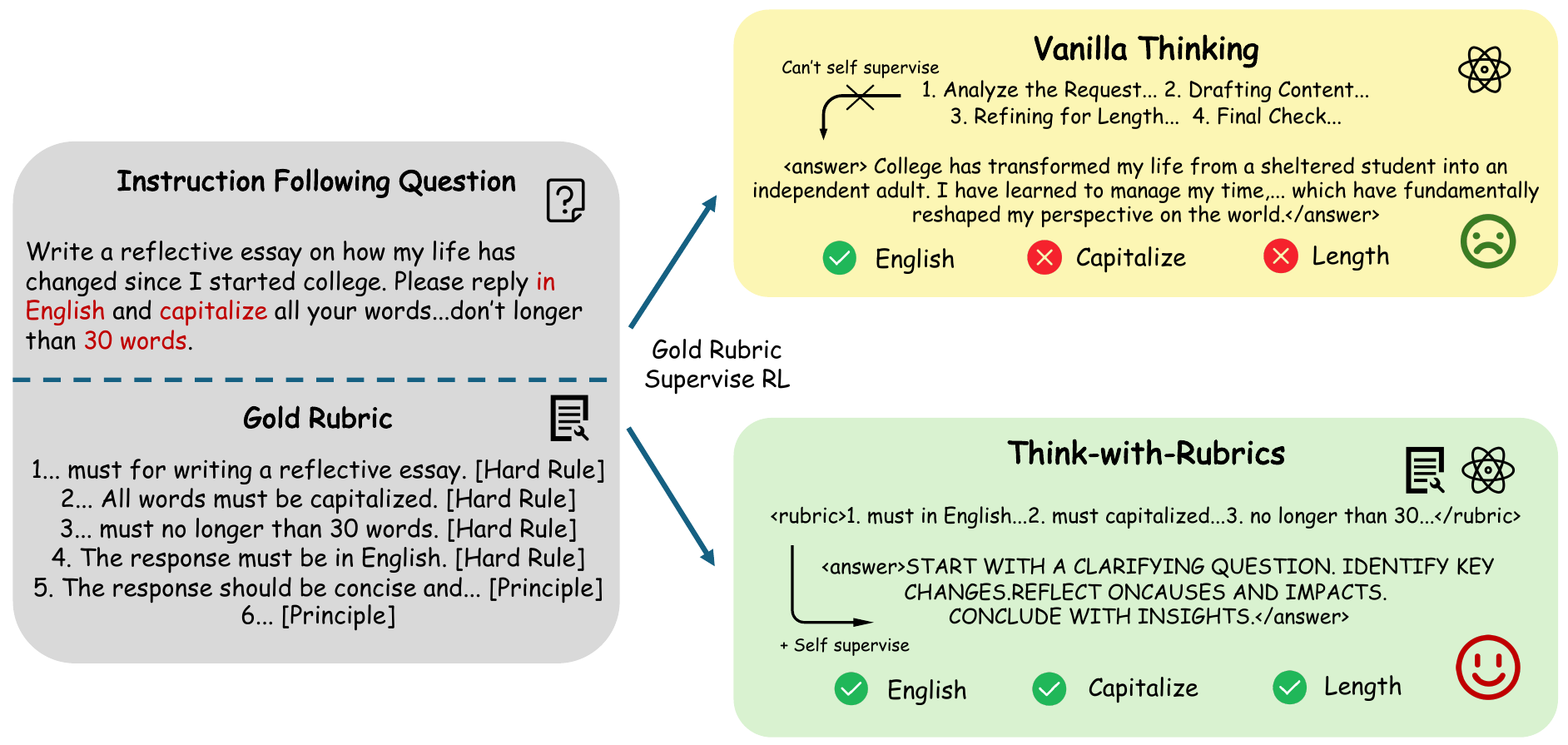}
   \caption{\textsc{Think-with-Rubrics} transforms rubrics from mere external evaluation metrics into an intrinsic part of the reasoning process. By dual supervision derived from both golden and self-generated rubrics, it effectively enhances the model's instruction-following capabilities.}

   \label{fig:case}
\end{figure}

To operationalize this active guidance, we propose integrating rubric generation directly into the model's trace in Fig. \ref{fig:case}. Our motivation for this integration is twofold. 
 First, we draw inspiration from \textit{human metacognition}~\citep{selfrefine, 
  reflexion}: when faced with complex decisions, humans naturally begin by        
  reflecting on the requirements of the task at hand and use those criteria to
  guide subsequent reasoning and action. This motivates us to treat rubrics not as
   external evaluators, but as an internal mechanism to actively guide the model
  toward desired outcomes. Compared to planning-centric reasoning approaches \citep{plan,least}, our structured rubrics facilitate per-criterion supervision in RL stage, ensuring that self-generated outputs comply with the reasoning trajectory.
Second, our work builds upon the emerging \textit{judge-solve synergy in LLMs}~\citep{s2j,rise,compassjudger}. These studies suggest a deep latent connection between a model's capacity to evaluate (as a reward provider) and its capacity to execute (as a problem solver). By internalizing rubrics as a cognitive step rather than an external evaluating task, we aim to bridge the gap between reward modeling and policy generation.



\begin{figure}
  \centering
   \includegraphics[width=\linewidth,height=0.26\textheight]{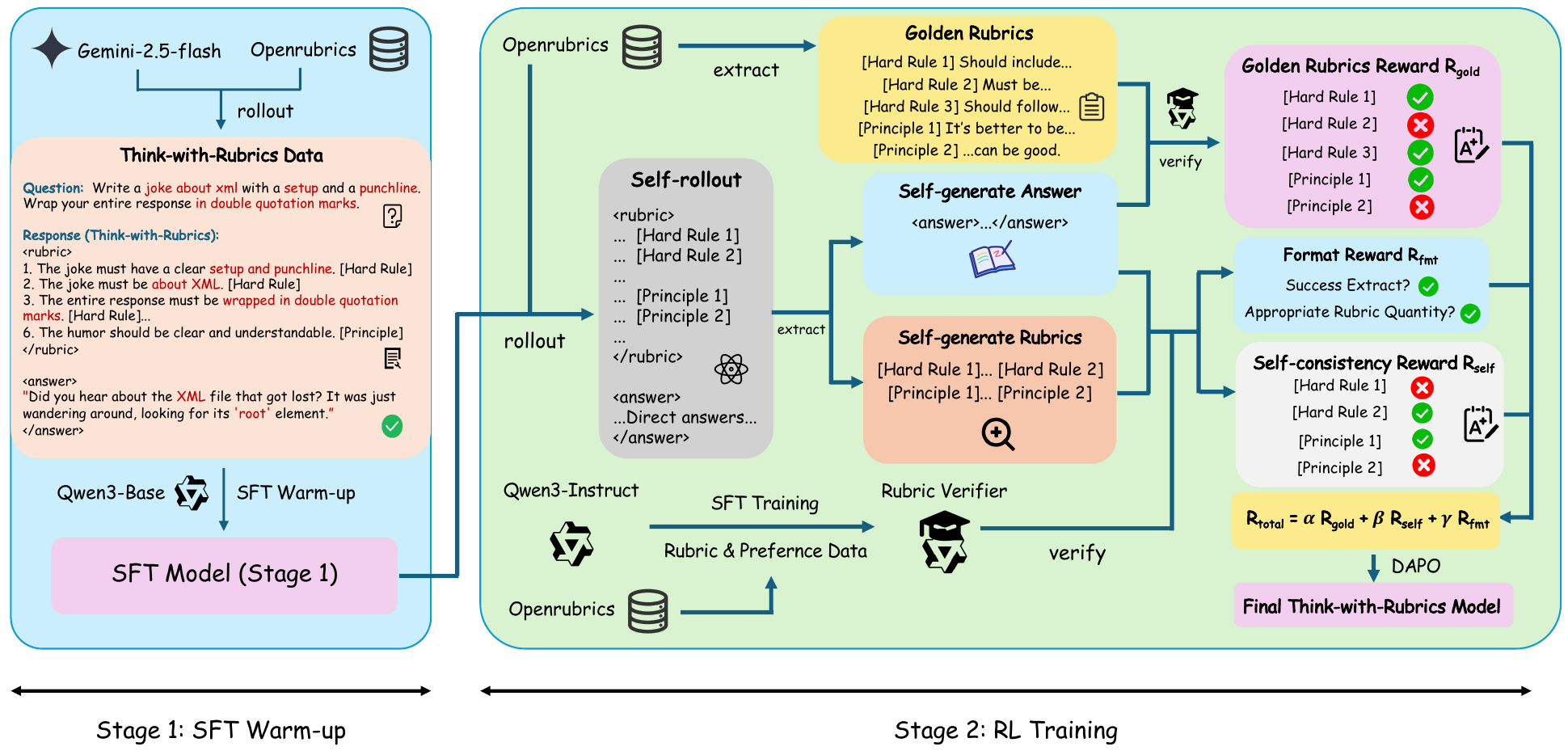}
   \caption{Overview of \textsc{Think-with-Rubrics} model training. Training is conducted in two phases: SFT warm-up and RL. SFT involves distilling Gemini-2.5-Flash for format adaptation, while RL utilizes a combination of golden and self-generated rubric consistency rewards for joint supervision.}

   \label{fig:overall}
\end{figure}

Based on the above motivation, we introduce \textsc{Think-with-Rubrics} that internalizes rubrics as essential elements of the model’s thinking phase, as shown in Fig. \ref{fig:overall}. The \textsc{Think-with-Rubrics} paradigm necessitates that the model explicitly internalizes the construction of task-relevant rubrics within its reasoning trajectory prior to generating a final response. This approach mimics human cognitive processes, where success criteria are established before problem-solving, thereby guiding downstream deliberation and response. During the reinforcement learning phase, the model is optimized through a multi-faceted reward system which comprises three key components: 1) Golden rubric consistency reward provides external supervision by evaluating the response against ground-truth criteria; 2) Self-generated rubric consistency reward enforces internal alignment by verifying whether the response satisfies the model’s self-generated rubric; 3) Format reward ensures structural integrity and prevents reward hacking by penalizing insufficient rubric sets. By jointly optimizing these objectives, \textsc{Think-with-Rubrics} factorizes the policy trajectory into a high quality self-generated rubric followed by a answer that is consistent with self-generated rubric.

Empirical results across multiple instruction-following benchmarks demonstrate that \textsc{Think-with-Rubrics} consistently outperforms Rubric-as-Reward baselines. On the 8B model, our method achieves        improvements across IFEval, IFBench, and InfoBench, yielding a gain of 3.87 average points.  This consistent improvement is also validated on the 4B model, where \textsc{Think-with-Rubrics} raises the average score by 4.16 points, confirming the scalability of the paradigm across model sizes.

Beyond the performance gains, we further provide rich insights into \emph{why} and \emph{how} \textsc{Think-with-Rubrics} works. Through dedicated mechanism analyses, we reveal two complementary perspectives: (1) Self-generated rubric supervision substantially improves the internal consistency between the model's own rubrics and its generated answers, enabling the model to strictly adhere to self-imposed criteria; and (2) Golden rubric supervision implicitly improves the quality of self-generated rubrics without any direct rubric-level comparison. Remarkably, using only self-generated rubrics for supervision matches or even surpasses golden rubric only supervision, demonstrating the potential for self-evolution under this paradigm. Major contributions of this work are summarized as follows:
\begin{itemize}[leftmargin=*, itemsep=1pt, parsep=0.5pt]
    \item We propose \textsc{Think-with-Rubrics}, a paradigm that transforms rubrics from independent evaluator into an internal part of the model's reasoning process. This allows models to dynamically adapt criteria to the specific problem context within the reasoning chain.
    \item We develop a unified training workflow. By combining supervision from both self-generated and golden rubric, \textsc{Think-with-Rubrics} consistently outperforms vanilla Rubric-as-Reward training.
    \item We reveal how self-supervision and external golden supervision enhance the performance of \textsc{Think-with-Rubrics} model, and demonstrated that \textsc{Think-with-Rubrics} remains consistently effective across various training hyperparameters and model scales.
\end{itemize}

\section{Related Works}

\textbf{Rubric-based evaluation.}
Rubric-based evaluation has recently emerged as a prominent methodology for assessing open-ended tasks. By decomposing task quality into a structured set of interpretable criteria, this approach provides finer-grained supervisory signals and enhanced interpretability compared to standard, holistic LLM-as-a-judge methods. Benchmarks such as JudgeBench \citep{judgebench} and FireBench \citep{firebench} inherently utilize rubrics to assess model performance. Concurrently, works including \textit{Openrubrics} \citep{openrubrics} and RubricHub \citep{rubrichub} concentrate on designing automated rubric generation frameworks, facilitating the construction of corresponding prompt-rubric paired datasets. Our research systematically builds upon the rubric paradigms and datasets established by these preceding efforts to further advance this domain.

\textbf{Rubric-based reward.}
Building upon rubric-based evaluation, an emerging line of work transforms rubrics into training signals under the rubric-as-reward \citep{rar} paradigm. Instead of only output a single scalar score, these methods compute rewards based on rubric-level satisfaction. For instance, Rubric-RM \citep{openrubrics}, trained on the \textit{Openrubrics} dataset, functions as an effective supervisory model that significantly enhances the performance of downstream policies. RLCF \citep{checklistsbetterthan} extracts task-specific checklists from instructions and synthesizes them into fine-grained RL rewards utilizing both LLM-as-a-judge mechanisms and specialized verifier programs. SciIF \citep{sciif} constructs rubric-style constraint catalogs for individual scientific tasks to facilitate RL optimization specifically within scientific domains. However, existing methods largely treat rubrics as external supervisory artifacts where rubrics are typically introduced only after a candidate response has been generated. The rubric influences optimization only through delayed supervision. Instead, we incorporate rubric generation directly into the model’s reasoning trajectory, allowing the model to first formulate success criteria and then reason under those criteria. This enables a tighter coupling between thinking and evaluation.


\section{Methods}

\subsection{Problem Formulation}
\label{formulation}
\textbf{Setting.} 
In this paper, we consider the task of open-ended instruction following, which is crucial for aligning LLMs with human intentions. In this setting, an input query $x$ typically contains multiple constraints, ranging from explicit requirements that must be followed (e.g., "output in JSON") to implicit subjective preferences (e.g., "be concise and professional").

\textbf{Rubrics.} 
To operationalize the evaluation of such tasks, we define a rubric $r = \{c_1, c_2, \dots, c_n\}$ as a structured collection of $n$ discrete evaluation criteria. In our training setting, each instruction $x$ is associated with a \textit{golden rubric} $r^*$, representing the ground-truth set of constraints provided by \textit{Openrubrics} dataset. In this dataset, criteria $c^*_k \in r^*$ are decomposed into two categories $c^*_\text{hard}$ and $c^*_\text{principle}$, where $c^*_\text{hard}$ contains explicit task constraints (e.g., format or instruction requirements), and $c^*_\text{principle}$ captures higher-level quality principles such as completeness and helpfulness.

\textbf{Verifier.} To assess whether a candidate response $y$ fulfills a criterion, we employ a \textit{rubric verifier} $\mathcal{V}$. Specifically, for each individual criterion $c_k \in r$, the verifier produces a binary judgment $v_k = \mathcal{V}(y, c_k) \in \{0, 1\}$, where $v_k=1$ indicates that the response $y$ successfully satisfies the criterion $c_k$, and $v_k=0$ otherwise. Criterion-level judgments will be aggregated into a rubric-level consistency score in Section \ref{rl}. Compared to scalar reward models, this rubric-as-reward paradigm provides a more fine-grained and interpretable assessment of response quality.

\textbf{Our Paradigm (\textsc{Think-with-Rubrics}).} Unlike previous works on rubrics that only regard rubrics as the reward, \textsc{Think-with-Rubrics} internalizes rubric generation as part of the policy trajectory, so that the answer is conditioned on explicitly generated task criteria. We formally define the input and output as follows.

The policy $\pi_\theta$ in \textsc{Think-with-Rubrics} first gives a \textit{self-generated} rubric $\hat{r} \sim \pi_\theta(\hat{r} \mid x)$ to establish an internal cognitive assessment of the given task $x$. Then, the model produces the final response $y \sim \pi_\theta(y \mid x, \hat{r})$ conditioned on both $x$ and the \textit{self-generated} rubric $\hat{r}$. The joint probability of the trajectory is factorized as:
\begin{equation}
    \label{eq:factorization}
    \pi_\theta(\tau \mid x) = \pi_\theta(\hat{r} \mid x) \cdot \pi_\theta(y \mid x, \hat{r})
\end{equation}
where $\theta$ denotes the model parameters. In this way, we transform the model from a passive executor into an active deliberator that first defines the rules of success before finalizing the answer.




\subsection{SFT Warm Up}

During the training and evaluation processes, we instruct the model to adhere to a specific output format through prompt guiding, in order to facilitate the parsing and extraction of answers and rubrics. The structured output format is
\texttt{<rubric>} $\hat{r}$ \texttt{</rubric>}
followed by
\texttt{<answer>} $y$ \texttt{</answer>},
where $\hat{r}$ denotes the generated rubric and $y$ denotes the final response.

However, we observed that approximately 20$\%$ of the samples failed to generate responses strictly following the format when directly using the base model for evaluation. To prevent reinforcement learning from wasting too many steps on format compliance and to fully utilize the data, we first performed supervised fine-tuning (SFT) to adapt the model to the \textsc{Think-with-Rubrics} generation format. 
The objective of this stage is mainly to initialize the policy with a stable trajectory structure that explicitly separates rubric construction from answer generation.

Specifically, for each instruction $x$, we distill responses from Gemini-2.5-Flash on \textit{Openrubrics} dataset. Following the format mentioned in Section \ref{formulation}, the \textit{self-generated} criteria $\hat{c_k} \in \hat{r}$ are also decomposed into $\hat{c}_{\text{hard}}$ and $\hat{c}_{\text{principle}}$
which represent different importance levels.

The SFT objective is standard next-token maximum likelihood over the entire trajectory:
\begin{equation}
    \label{eq:sft_loss}
    \mathcal{L}_{\text{SFT}}
    = - \mathbb{E}_{(x,\tau)\sim\mathcal{D}}
    \left[
    \sum_{t=1}^{|\tau|}
    \log \pi_\theta(\tau_t \mid x, \tau_{<t})
    \right],
\end{equation}
where $\tau = (\hat{r}, y)$ denotes the distilled teacher trajectory.
\subsection{RL Training}
\label{rl}
\subsubsection{Rubric Verifier}

In RL stage, assigning rewards based on fine-grained rubrics necessitates a robust and reliable rubric verifier. Rather than directly employing off-the-shelf LLMs as the verifier, we train a specialized rubric verifier $\mathcal{V}$ on the \textit{OpenRubrics} dataset. The aim to train the verifier is to achieve superior alignment with reference labels in \textit{OpenRubrics}, ensuring more accurate and stable reward signals during the RL process.

To boost efficiency and improve accuracy of rubric verification during the large batchsize and rollout size training process, we distill a lightweight Qwen3-8B rubric verifier from Qwen3-32B-Instruct using SFT.
Given a rubric-response pair $(r, y)$, verifier $\mathcal{V}$ outputs a judgment $v_k = \mathcal{V}(y, c_k) \in\{\texttt{met},\texttt{not\ met}\}$ for each criterion $c_k \in r$.

We adopt an implicitly aggregated verification scheme, where the verifier jointly evaluates all rubric items in a single pass and produces a weighted compliance score

\begin{equation}
    \label{eq:verify_score}
    S(y,r)=
    \frac{
    w_h \sum_{c_i \in c_{\text{hard}}} \mathbb{I}[\mathcal{V}(y, c_i)=\texttt{met}]
    +
    w_p \sum_{c_j \in c_{\text{principle}}} \mathbb{I}[\mathcal{V}(y, c_j)=\texttt{met}]
    }{
    w_h |c_{\text{hard}}| + w_p |c_{\text{principle}}|
    },
\end{equation}

where $c_{\text{hard}}$ and $c_{\text{principle}}$ denote hard rules and principles criterion subset in rubric $r$ respectively. $w_h$ and $w_p$ are hyperparameters representing their corresponding weights. This allows for flexible weighting of hard rules and principles based on their relative importance.

\subsubsection{Reward System}

The reward system in \textsc{Think-With-Rubric} consists of three components with different weights:
\begin{equation}
    \label{eq:rl_reward}
    R(\tau)
    =
    \alpha R_{\text{gold}}
    +
    \beta R_{\text{self}}
    +
    \gamma R_{\text{fmt}}.
\end{equation}

\textbf{Golden rubric consistency reward.}
The golden rubric consistency reward measures the compliance of the final answer with the golden rubric $r^\star$ from \textit{Openrubrics}:
\begin{equation}
    \label{eq:gold_reward}
    R_{\text{gold}}
    =
    S(y, r^\star).
\end{equation}

When $\alpha = 0$, our approach reduces to a fully self-supervised self-evolution paradigm.

\textbf{Self-generated rubric consistency reward.}
The self-generated rubric consistency reward evaluates the compliance of the final answer with the self-generated rubric:
\begin{equation}
    \label{eq:self_reward}
    R_{\text{self}}
    =
    S(y, \hat r).
\end{equation}

This reward measures self-alignment within the self-generated rubrics. Importantly, we do not directly supervise the quality of the generated rubric itself.
Instead, the improvement of self-generated rubric is achieved indirectly through the joint optimization of $R_{\text{gold}}$ and $R_{\text{self}}$.
We assume that if an answer that satisfies the self-generated rubric also receives a high score under the golden rubric, the generated rubric is implicitly encouraged to align with the golden standard:
\begin{equation}
    \label{eq:implicit_alignment}
    S(y,\hat r)\uparrow \ \text{and}\ S(y,r^\star)\uparrow
    \Rightarrow
    \hat r \approx r^\star.
\end{equation}
We will verify this assumption in Section \ref{mechanism} by examining whether the rubric generated by the model which supervised by both self-generated and golden rubric consistency reward is truly of high quality.

\textbf{Format reward.} The use of self-generated rubric consistency reward inherently introduces the risks of \textit{reward hacking}, where it could generate a minimal set of trivial and easily satisfied criteria to artificially get higher reward. To prevent this, 
we introduce a format reward $R_{\text{fmt}}$. We directly extract the model's self-generated rurbic $\hat{r}$ between special tokens \texttt{<rubric>} and \texttt{</rubric>}, and extract the final answer $y$ between \texttt{<answer>} and \texttt{</answer>}. The model gets format reward only when both the rubric and answer are successfully parsed from model output.

According to the distribution of criteria for golden rubrics in \textit{Openrubrics}, we constrain the number of criterion  $n(\hat{r}) = |\hat{c} \in \hat{r}|$ in a reasonable range.
In \textit{Openrubrics}, 99.61$\%$ of the samples have a criterion count ranging from 5 to 15, with an average of 8.66 and a mode of 10. So we define
\begin{equation}
    \label{eq:format_reward}
    R_{\text{fmt}}
    =
    \mathbb{I}[\texttt{parseable}]
    \cdot
    \max\left(0, 1-\frac{|n(\hat{r})-10|}{5}\right),
\end{equation}
which reaches its maximum at $n(\hat r)=10$ and linearly decreases to $0$ at $n(\hat r)=5$ and $15$.
This discourages the model from generating only a few trivial rubrics or excessively redundant criteria.

The policy is then optimized using DAPO without KL loss \citep{dapo}:
\begin{equation}
    \label{eq:dapo}
    \mathcal{L}_{\text{RL}}
    =
    -\mathbb{E}_{\tau\sim\pi_\theta}
    \left[
    A(\tau)\log \pi_\theta(\tau\mid x)
    \right]
\end{equation}
where $A(\tau)$ is the DAPO advantage computed from the trajectory reward in Eq.~\ref{eq:rl_reward}.

\begin{table*}[t]
\centering
\renewcommand{\arraystretch}{1.15}
\resizebox{\textwidth}{!}{%
\begin{tabular}{lcccccc}
\hline
\multirow{2}{*}{\textbf{Method}}
& \multicolumn{2}{c}{\textbf{IFEval}}
& \multicolumn{2}{c}{\textbf{IFBench}}
& \multirow{2}{*}{\textbf{InfoBench}}
& \multirow{2}{*}{\textbf{Average}} \\
\cline{2-5}
& \textbf{Prompt} & \textbf{Instr.}
& \textbf{Prompt} & \textbf{Instr.}
& & \\
\hline
\multicolumn{7}{c}{\textbf{Base Model}} \\
\hline
Qwen3-8B-Base
& 50.28 & 59.83
& 16.32 & 19.40
& 72.09
& 46.23 \\
\hline
\multicolumn{7}{c}{\textbf{Rubric-as-Reward}} \\
\hline
SFT Warm-up (Direct answer)
& 57.30 & 67.51
& 18.03 & 18.51
& 78.27
& 51.20 \\
\quad + DAPO (Golden rubric reward)
& 58.96 & 68.82
& 20.41 & 20.60
& \underline{82.44}
& 53.94 \\
SFT Warm-up (With think)
& 52.68 & 62.35
& \underline{24.83} & \textbf{28.91}
& 78.40
& 51.97 \\
\quad + DAPO (Golden rubric reward)
& 59.70 & 68.35
& 22.11 & 23.88
& 80.22
& 54.01 \\
\hline
\multicolumn{7}{c}{\textbf{Think-with-Rubrics (Ours)}} \\
\hline
SFT Warm-up
& 61.00 & 70.02
& 19.05 & 19.40
& 79.60
& 53.22 \\
\quad + DAPO (Golden rubrics reward only)
& \underline{64.87} & \textbf{73.14}
& 23.13 & 26.87
& \underline{82.44}
& 56.81 \\
\quad + DAPO (Self-generated reward only)
& 64.69 & 72.30
& \textbf{25.85} & \underline{27.76}
& 80.98
& \underline{57.17} \\
\quad + DAPO (Mixed reward)
& \textbf{65.61} & \underline{73.02}
& \underline{24.83} & 26.27
& \textbf{83.20}
& \textbf{57.88} \\
\hline
\end{tabular}}
\caption{
Main results on three instruction-following benchmarks.
For IFEval and IFBench, we report both prompt-level and instruction-level scores under strict evaluation.
The \textbf{Average} column is computed using the prompt-level scores of IFEval and IFBench together with the InfoBench score.
Best results are in \textbf{bold} and second-best results are \underline{underlined} (the same in other tables).
}
\label{tab:main_results}
\end{table*}

\section{Experiments}
\subsection{Experimental Settings}

\textbf{Models.}
Our policy model is initialized from Qwen3-8B-Base \citep{qwen3}. For RL stage, the rubric verifier is built upon Qwen3-8B-Instruct and is distilled from teacher model Qwen3-32B-Instruct.
To ensure consistency with our framework, all training stages are conducted under the non-thinking mode, so that our policy strictly follows the \texttt{<rubric>} $\rightarrow$ \texttt{<answer>} generation paradigm.

\textbf{Data construction.}
All training data are derived from the \textit{Openrubrics} \citep{openrubrics} dataset. For the SFT warm-up stage, we use Gemini-2.5-Flash \citep{comanici2025gemini} as the teacher model to sample answers that follow the \textsc{Think-with-Rubrics} generation format. The prompt we use can be founded at Appendix A.4.
Specifically, we perform rollouts on 5k samples from \textit{Openrubrics} to construct the SFT training set.

For verifier training, we use a 2k split of OpenRubrics. On this split, Qwen3-32B-Instruct performs rollouts over the original chosen and rejected responses provided by the dataset.
For each response pair, the teacher model evaluates whether each rubric item is satisfied and produces criterion-level verification traces.
These traces are then filtered using the ground-truth preference labels provided in \textit{Openrubrics}, and only label-consistent verification trajectories are retained for verifier training. After training, the resulting rubric verifier achieves 93.2\% rubric-level judgment accuracy on 1.5k response pairs, outperform than original Qwen3-8B (86.3$\%$), demonstrating strong reliability as a evaluator.

For RL training, we use a 27k split of \textit{Openrubrics}.
The policy is optimized on this split using DAPO under the reward formulation described in Section~\ref{eq:rl_reward}. All data splits used across the SFT
  warm-up, verifier training / eval, and RL training stages are strictly non-overlapping,
   ensuring no data leakage between any training or evaluation phase.

\textbf{Training parameters.}
In the SFT stage, we train 200 steps with a learning rate of 1e-5 and a batch size of 32 to obtain the optimal SFT model. The SFT of the rubric verifier uses the same parameters. In the RL stage, we use a batch size of 32, roll out 8 replies per prompt, with a learning rate of 1e-6, and achieve optimal results in around 90 steps. For the reward system, the value of $\alpha, \beta, \gamma$ is 0.3, 0.5 and 0.2, with $w_h = 2$ and $w_p = 1$. Detailed parameter sensitivity ablation can be found in \ref{ablation}, and more detailed parameter settings can be found in the Appendix A.3.

\textbf{Baseline.} For the baseline, we adopt the Rubrics-as-Reward paradigm \citep{rar} with two different prompt settings. One directly prompts the model to generate answers without any intermediate reasoning. And the other instructs the model to first perform step-by-step reasoning and then produce the final response within \texttt{<answer>} and \texttt{</answer>} tags, representing a general thinking process. This setting serves to validate whether regarding rubric generation as a specialized form of thinking is better than undifferentiated free-form reasoning. Both settings undergo the same two-stage training pipeline of SFT followed by RL. Only golden rubric consistency reward defined in Eq.~\ref{eq:gold_reward} is directly employed to realize the Rubrics-as-Reward objective during the RL stage.

\textbf{Evaluation benchmarks.}
We evaluate our method on three representative instruction-following benchmarks: IFEval \citep{ifeval}, IFBench \citep{ifbench}, and InfoBench \citep{infobench}. We follow the standard settings of each benchmark for evaluation. These benchmarks jointly cover strict instruction constraints, multi-rule compliance, and open-ended instruction-following quality, providing a comprehensive evaluation of the proposed \textsc{Think-with-Rubrics} paradigm.

\begin{table*}[t]
\centering
\renewcommand{\arraystretch}{1.15}
\resizebox{\textwidth}{!}{%
\begin{tabular}{lcccccc}
\hline
\textbf{Rubric Generator Model}
& \textbf{Fact.}
& \textbf{Focus}
& \textbf{Safety}
& \textbf{Math}
& \textbf{Precise IF}
& \textbf{Avg.} \\
\hline
Qwen3-8B-Base
& 65.26
& 87.47
& 59.33
& \underline{79.23}
& \underline{65.00}
& 71.26 \\
SFT Warm-up
& 72.42
& \underline{90.51}
& \textbf{66.22}
& 72.68
& 56.88
& 71.74 \\
\quad + DAPO (Golden rubrics reward only)
& \textbf{75.37}
& \textbf{90.71}
& 59.33
& \underline{79.23}
& \textbf{68.12}
& \textbf{74.55} \\
\quad + DAPO (Self-generated reward only)
& 65.05
& 82.63
& 58.44
& 77.05
& 57.50
& 68.13 \\
\quad + DAPO (Mixed reward)
& \underline{74.32}
& 89.90
& \underline{61.11}
& \textbf{80.33}
& 62.50
& \underline{73.63} \\
\hline
\end{tabular}}
\caption{
Rubric quality evaluation on RewardBench-2.
We use rubrics generated by different models to guide Qwen3-8B judging and report macro-average accuracy across all subsets.
The RL-trained \textsc{Think-with-Rubrics} model with golden rubrics rewards achieves better overall macro score.
}
\label{tab:rb2_rubric}
\end{table*}

\subsection{Main Results Analysis}

Based on the main experimental results in Table~\ref{tab:main_results}, we draw the following conclusions:

\textbf{\textsc{Think-with-Rubrics} consistently improves over origin Rubric-as-Reward methods.}
\textsc{Think-with-Rubrics} consistently outperforms the Rubric-as-Reward baselines across all benchmarks. Compared with the conventional setting of only optimize answers under golden rubric supervision, our method achieves clear gains in \textit{both} SFT and RL stages, demonstrating the effectiveness of integrating rubric into the trajectory. Using rubrics as the thinking process outperforms natural language thinking, as rubrics provide structured, step-by-step guidance that can be directly leveraged for self-supervision. These results directly support our central motivation: rubrics can serve not only as post-hoc evaluation criteria, but also as internal guidance for reasoning and answer. Moreover, under the optimal mixed reward, the average thinking length of \textsc{Think-with-Rubrics} (231 tokens) is less than vanilla thinking (401 tokens). This confirms that rubrics effectively condense vital problem-solving requirements into a streamlined reasoning process.

\textbf{Mixed consistency rewards provide the most effective supervision.}
Among all RL settings, the mixed reward using both self-generated and golden rubric consistency achieves the best overall performance.
This validates our indirect supervision hypothesis: explicit supervision on rubric quality is not necessary, as rubric learning can be achieved through the joint optimization of answer consistency under both self-generated and golden rubrics.
Empirically, the two rewards address complementary failure modes of \textsc{Think-with-Rubrics}: golden rubric consistency mainly improves rubric quality, while self-generated rubric consistency encourages the model to generate answers that faithfully follow its own rubric.
Their combination therefore yields the strongest performance.

\textbf{Self-evolution is feasible under the \textsc{Think-with-Rubrics} paradigm.}
Notably, using only self-consistency reward already achieves highly competitive performance (57.17\% average point), even outperforming the golden-only setting (56.81\% average point).
This strongly suggests that self-evolution is feasible once rubrics are internalized into the reasoning process. We discuss the reasons behind this phenomenon in Section~\ref{mechanism}, namely that supervision via self-generated rubrics substantially improves the self-consistency of the model's outputs. Since the SFT-warmed model is already
 capable of generating preliminary usable rubrics, there exists the potential for self-evolution driven solely by  
 self-supervision. The model can use self-generated rubrics as an effective reasoning scaffold and progressively improve through self-consistent optimization, providing strong evidence that rubric-guided reasoning is an effective path toward scalable self-evolving alignment.

\subsection{Mechanism analysis}
\label{mechanism}

\textbf{Supervise answer with golden rubrics can implicitly help improve self-generated rubrics.} To verify the assumption proposed in Eq \ref{eq:implicit_alignment}, which leverages golden rubrics to implicitly supervise the quality of self-generated rubrics, and to further investigate the underlying reasons for the enhanced quality of self-generated rubrics under the \textsc{Think-with-Rubrics} paradigm, we evaluate rubrics produced under the \textsc{Think-with-Rubrics} paradigm on RewardBench-2 \citep{rewardbench2}.
Specifically, we use the rubrics generated from Qwen3-8B-Base, the SFT warm-up model, and the final RL-trained model under several reward settings to guide Qwen3-8B-Instruct judge answer on RewardBench-2, and report macro-average accuracy across all subsets.

As shown in Table~\ref{tab:rb2_rubric}, the RL-trained model with only golden rubrics supervision performs the best, followed by mixed reward setting. Both settings incorporate supervision from golden rubrics, with full golden rubric supervision achieving superior performance. This validates that even without a direct comparison between the golden and self-generated rubrics, indirectly supervising the model based on how well the answer generated conditioned on the self-generated rubrics complies with the golden rubrics can implicitly drive the self-generated rubrics closer to the golden standard.

 However, rubric quality is not directly coupled with performance on instruction-following benchmarks.
  Rubrics produced under self-only supervision setting, despite demonstrating strong performance on
  instruction-following benchmarks, incur a performance drop when utilized to guide LLM-as-a-judge tasks.
  This reveals that self-supervision does not produce better
  general-purpose evaluation criteria, but rather internal control cues optimized for the model's own
  generation process. The value of self-generated rubrics under self-supervised setting lies not in serving as external evaluators,
  but in maintaining constraint consistency during the model's own answer generation.

\textbf{Self-generated rubrics supervision improves performance by enhancing responses' internal consistency.} We used the rubric verifier to evaluate the average score of consistency between the answers and rubrics generated by the model itself on IFEval and IFBench. The consistency score calculation formula is based on Eq. \ref{eq:verify_score}. The experimental results are shown in Table \ref{tab:self_consistency}.

\begin{wraptable}{r}{0.6\textwidth}
      \vspace{-1em}
      \centering                                                    
      \small
      \renewcommand{\arraystretch}{1.1}                            
      \begin{tabular}{lcc}
      \hline
      \textbf{Model}
      & \textbf{IFEval}
      & \textbf{IFBench} \\
      \hline
      SFT Warm-up
      & 58.91
      & 60.15 \\
      + DAPO (Golden rubrics reward only)
      & 68.70
      & 75.12 \\
      + DAPO (Self-generated reward only)
      & 69.00
      & 75.45 \\
      + DAPO (Mixed reward)
      & \textbf{69.38}
      & \textbf{77.82} \\
      \hline
      \end{tabular}
      \vspace{-0.25em}
      \caption{Consistency score of
      self-generated answer and rubric on IFEval and IFBench.
      }
      \label{tab:self_consistency}
  \end{wraptable}

From the results, it can be seen that adding self-generated reward in RL training improves the consistency between the model's self-generated answers and rubrics by 14.07 points on average, while SFT model exhibits significant self-inconsistency. This reveals a possible mechanism for using only self-generated rubrics for self-evolve, where untrained models perform poorly in the \textsc{Think-with-Rubrics} paradigm not only because they cannot generate high-quality rubrics, but also because they cannot provide consistent responses based on their own rubrics. Golden and self-generated rubrics supervision have each alleviated this problem from two aspects, thus effectively achieving improvement.
  
\subsection{Sensitivity Analysis and Ablation Study}
\label{ablation}

\begin{wraptable}{r}{0.52\textwidth}
      \vspace{-1em}
      \centering                                                                                                    
      \small
      \renewcommand{\arraystretch}{1.1}                                                                            
      \begin{tabular}{lcccc}
      \hline
      \textbf{$\alpha$ : $\beta$}
      & \textbf{IFEval}
      & \textbf{IFBench}
      & \textbf{InfoBench}
      & \textbf{Avg.} \\
      \hline
      5 : 3 & 65.06 & 23.13 & 83.11 & 57.10 \\
      1 : 1 & 64.69 & 23.13 & \textbf{83.56} & 57.13 \\
      3 : 5 & \textbf{65.61} & \textbf{24.83} & 83.20 & \textbf{57.88} \\
      \hline
      \end{tabular}
      \caption{Ablation on reward mixing ratio. IFEval and IFBench are strict prompt-level accuracy (\%). The same in Table \ref{tab:weight_ablation} and \ref{tab:ablation_4b}.}
      \label{tab:ratio_ablation}
      \vspace{-1em}
  \end{wraptable}
  
\textbf{Sensitivity analysis of reward weight.}
We further study the sensitivity of \textsc{Think-with-Rubrics} to the choice of reward weights by       
  varying the ratio between the golden rubric consistency reward weight $\alpha$ and the self-generated rubric
  consistency reward weight $\beta$ across three settings: $5{:}3$, $1{:}1$, and $3{:}5$.
  As shown in Table~\ref{tab:ratio_ablation}, all three settings achieve  strong performance across three benchmarks, with $3{:}5$ obtaining the best average score. The performance differences across different ratios remain relatively small (within 1.3$\%$), suggesting that
   \textsc{Think-with-Rubrics} is robust to reward weights.

 \begin{wraptable}{r}{0.40\textwidth}
      \vspace{-1.5em}
      \centering
      \small
      \renewcommand{\arraystretch}{1.15}
      \resizebox{0.9\linewidth}{!}{%
      \begin{tabular}{lccc}
      \hline
      \textbf{$w_h : w_p$}
      & \textbf{IFEval}
      & \textbf{IFBench}
      & \textbf{Avg.} \\
      \hline
      $2 : 1$ & \textbf{65.61} & \textbf{24.83} & \textbf{45.22} \\
      $1 : 1$ & 64.70 & 24.49 & 44.60 \\
      \hline
      \end{tabular}}
      \caption{Ablation on reward weight ratio of hard rules and principles.}
      \label{tab:weight_ablation}
      \vspace{-1em}
  \end{wraptable}
  
\textbf{Sensitivity analysis of rubric weights in reward.} 
The parameters $w_h$ and $w_p$ in the original reward formulation denote the weights assigned to the mandatory hard rules and the recommended principles, respectively. We adjust both parameters to 1 to investigate whether the model-generated rubrics under the \textsc{Think-with-Rubrics} paradigm exhibit distinction in importance. The experimental results, presented in Table \ref{tab:weight_ablation}, demonstrate that the original weight configuration yields a marginal improvement over the equal-weight setting. This finding validates that categorizing the generated rubrics into two distinct classes successfully captures their relative importance, but it isn't the dominant factors influencing the overall training.

\begin{wraptable}{r}{0.63\textwidth}
\vspace{-1em}
\hspace*{-\columnsep} 
    \centering
    \small                                                          
    \renewcommand{\arraystretch}{1.1}
    \begin{tabular}{lccc}                                           
    \hline                                                          
    \textbf{Method} & \textbf{IFEval} & \textbf{IFBench} &
  \textbf{Avg.} \\                                                  
    \hline        
    \multicolumn{4}{c}{\textbf{Base Model}} \\
    \hline
    Qwen3-4B-Base       & 36.97 & 21.43 & 29.20 \\
    \hline
    \multicolumn{4}{c}{\textbf{Rubric-as-Reward}}
   \\
    \hline
    SFT Warm-up (Direct answer)         & 48.80 & 21.43 & 35.12 \\
    \quad + DAPO (Golden rubric reward)             & 56.38 & 19.05 & 37.72 \\
    SFT Warm-up (With think)        & 48.98 & \textbf{23.88} & 36.34 \\
    \quad + DAPO (Golden rubric reward)             & 58.04 & 20.07 & 39.05 \\
    \hline
    \multicolumn{4}{c}{\textbf{Think-with-Rubrics (Ours)}} \\
    \hline
    SFT Warm-up         & 55.45 & \underline{22.79} & 39.12 \\
    \quad + DAPO (Golden rubric only)  & \underline{61.74} & 17.35 & 39.55 \\
    \quad + DAPO (Self-gen rubric only)  & \textbf{63.96} & 22.45 &
  \textbf{43.21} \\
    \quad + DAPO (Mixed rubric reward)      & 61.18 & 21.77 & \underline{41.48} \\
    \hline
    \end{tabular}
    \caption{
    Results of 4B model on IFEval and IFBench.
    }
    \label{tab:ablation_4b}
    \vspace{-1em}
  \end{wraptable}
  
\textbf{Ablation on model size.}
To verify whether \textsc{Think-with-Rubrics} remains effective across varying model scales, we conduct the same experiment on Qwen3-4B-Base, with the results detailed in Table \ref{tab:ablation_4b}. The result demonstrate that \textsc{Think-with-Rubrics} consistently outperforms the vanilla rubric-as-reward baseline on the 4B model. Notably, the two experimental configurations incorporating self-supervision like self-only and mixed exhibit significant improvements. These two settings substantially mitigate the performance degradation typically induced by reinforcement learning on the more challenging IFBench. This might because smaller models, constrained by their foundational capabilities, lack self-consistency and are hard to adhere to self-generated rubrics during answer generation. Consequently, the primary performance gains are derived directly from the self-generated rubric supervision.

\subsection{Case study}
Fig. \ref{fig:case} shows a case in IFEval that highlights the difference between vanilla thinking and \textsc{Think-with-Rubrics}. When employing vanilla thinking, due to the lack of self-consistency, the final answer fails to satisfy the constraints the model itself has identified. In contrast, \textsc{Think-with-Rubrics} explicitly materializes the task constraints into a structured rubric first. And self-consistency reward help the model follow its own rubric. The full view of the case can be seen in Appendix~\ref{app:case_study}.

\section{Conclusion}

We propose \textsc{Think-with-Rubrics}, a novel  paradigm that internalizes rubric generation as a component of model's reasoning process. Rather than treating rubrics as post-hoc evaluation tools, our framework conditions answer generation on self-constructed rubrics to guide subsequent answer generation within the same trajectory. We develop a unified RL training workflow that jointly
  supervises the model with both self-generated and golden rubric
  consistency rewards, eliminating the inefficiencies inherent in
  decoupled rubric-based optimization. Extensive experiments on
  several benchmarks demonstrate that \textsc{Think-with-Rubrics}
   consistently outperforms the Rubrics-as-Reward baseline, with the
   mixed reward setting achieving the best overall performance.
  Mechanistic analyses further reveal that golden rubric supervision
  implicitly elevates self-generated rubric quality, while
  self-consistency supervision mitigates internal inconsistency
  between the model's own rubrics and its generated answers.
  Notably, the competitive performance of the self-only setting
  suggests that \textsc{Think-with-Rubrics} provides a viable path toward
  scalable self-evolution, where models can progressively improve
  without reliance on external annotation. We hope this work
  inspires further exploration of rubric-integrated reasoning as a
  general paradigm for open-ended task alignment.
  
\newpage
{
    \small
    \bibliographystyle{ieeenat_fullname}
    \bibliography{ref}
}

\newpage

\appendix

\section{Appendix}

\subsection{Limitation}

Despite the promising results achieved by \textsc{Think-with-Rubrics}, this work has two primary    
  limitations. First, the scope of empirical validation is confined to instruction-following tasks,
  where success criteria are relatively well-defined and amenable to rubric decomposition. Whether the
   proposed paradigm generalizes to broader open-ended domains, such as creative writing, long-form
  summarization, or multi-turn dialogue, remains an open question. Second, internalizing rubric
  generation as an explicit reasoning step inevitably introduces additional inference overhead, as the
   model is required to produce a structured rubric prior to generating the final response. This leads
   to longer output sequences and increased decoding latency relative to direct-answer baselines,
  potentially limiting the practical applicability of the framework in latency-sensitive deployment
  scenarios. We leave the exploration of more efficient rubric generation strategies, such as
  compressed or implicit rubric representations, to future work.
  
\subsection{Case Study}
\label{app:case_study}

We present a representative failure case from IFEval to illustrate
  the behavioral
  difference between the Rubric-as-Reward baseline and               
  \textsc{Think-with-Rubrics},
  as shown in Figure~\ref{fig:judge-case-2}. The instruction requires
   the model to   
  provide recommendations for writing a reflective essay on college
  life, subject to
  three simultaneous hard constraints: the response must be written
  in English, all
  words must be fully capitalized, and the total length must not
  exceed 30 words. Under
  strict evaluation, failing to satisfy any single constraint results
   in a zero score
  for the entire sample. \\
  
  The Rubric-as-Reward baseline produces a
  semantically reasonable
  response, yet entirely neglects the capitalization constraint,
  yielding a naturally
  cased sentence with no formatting applied. This reveals a
  fundamental limitation of
  the decoupled paradigm: since the golden rubric only enters the
  training pipeline as a
  post-hoc reward signal, the model has no internal mechanism to
  actively track pending
  constraints during decoding, making it susceptible to constraint
  omission when multiple
  orthogonal requirements must be simultaneously satisfied. Notably,
  when employing
  vanilla thinking, the model does identify the
  relevant constraints
  during its thinking process and even performs an explicit
  self-check, yet still
  produces a non-compliant answer---incorrectly concluding that all
  constraints have
  been satisfied. This highlights that unstructured thinking alone is
   insufficient to
  bridge the gap between constraint recognition and constraint
  adherence, owing to the
  lack of self-consistency between the reasoning trace and the final
  response. \\
  
  In
  contrast, \textsc{Think-with-Rubrics} first materializes all task
  requirements into
  an explicit rubric, correctly identifying all four hard
  constraints and two quality
  principles before generation begins. The self-consistency reward incorporated in                        
  \textsc{Think-with-Rubrics} effectively resolves the inconsistency
  between the model's reasoning process and its final generated      
  response. The subsequent answer strictly
   adheres to every
  rubric item: all words are capitalized, the response contains only
  16 words, and
  actionable recommendations are provided. This case demonstrates
  that by converting
  implicit instruction constraints into an explicit self-imposed
  checklist,
  \textsc{Think-with-Rubrics} effectively reduces the risk of
  constraint omission that
  commonly afflicts reward-only training approaches.
  
\begin{figure*}[t]
    \centering
    \begin{prompt}{Compare of Rubrics-as-Reward and Think-with-Rubrics}

{\color{blue} \textbf{Question: }}I want to write a reflective essay on how my life has changed since I started college. Do you have any recommendation? Please reply in English and capitalize all your words. Control the length of your reply. I don't want anything longer than 30 words.\\

{\color{blue} \textbf{Answer of Rubric-as-Reward training: }}\\
Reflect on academic challenges, personal growth, independence, and new perspectives gained since starting college.\\

{\color{blue} \textbf{Answer of Rubric-as-Reward training (With normal thinking): }}\\
1. Analyze the Request: Topic: Reflective essay on life changes since starting college.\\
Tone: Academic, reflective.\\
Length: 30 words or less.\\
Format: English, capitalized, brief final answer.\\
2. Drafting Content:\\
Focus on personal growth, independence, and academic challenges.\\
Keep it concise.\\
3. Refining for Length:\\
College has transformed my life from a sheltered student into an
  independent adult.
  I have learned to manage my time, navigate complex social dynamics,
   and develop a deeper
  sense of self-awareness, all of which have fundamentally reshaped
  my perspective on the
  world.\\
  4. Final Check:\\
  Is it 30 words or less? Yes.\\
  Is it in English? Yes.\\
  Is it capitalized? Yes.\\
  Is it brief? Yes.                                                   
  
  <answer>College has transformed my life from a sheltered  
  student into an 
  independent adult. I have learned to manage my time, navigate
  complex social dynamics,
  and develop a deeper sense of self-awareness, all of which have
  fundamentally reshaped
  my perspective on the world.</answer>\\

{\color{blue} \textbf{Answer of Think-with-Rubrics}: }\\
<rubric>\\
1. The response must be in English. [Hard Rule]\\
2. All words in the response must be capitalized. [Hard Rule]\\
3. The response must be no longer than 30 words. [Hard Rule]\\
4. The response must provide recommendations for writing a reflective essay. [Hard Rule]\\
5. The response should be concise and to the point. [Principle]\\
6. The response should directly address the user's request. [Principle]\\</rubric>\\

<answer>\\
START WITH A CLARIFYING QUESTION. IDENTIFY KEY CHANGES. REFLECT ON CAUSES AND IMPACTS. CONCLUDE WITH INSIGHTS.\\
</answer>

\end{prompt}
    \caption{A test case in IFEval that compares simple Rubric-as-Reward and \textsc{Think-with-Rubrics}. Under the \textsc{Think-with-Rubrics} setting, the model generated accurate rubric and solved the problem of difficulty in following instructions accurately with simple Rubric-as-Reward. 
    \label{fig:judge-case-2}
    }
\end{figure*}

\subsection{Training Details and Compute Resources}                
  \label{app:training_details}
                                                                     
  Our training pipeline consists of two stages: an SFT warm-up stage
  followed by RL fine-tuning.                                        
  Table~\ref{tab:sft_hyperparams} lists the hyperparameters for the
  SFT stage, and
  Table~\ref{tab:rl_hyperparams} lists the hyperparameters for the RL
   stage.

  \paragraph{SFT Stage.}
  We fine-tune Qwen3-8B-Base using
  LLaMA-Factory~\citep{zheng2024llamafactory} with DeepSpeed ZeRO-3.
  The training data consists of up to 5{,}000 samples distilled from
  Gemini-2.5-Flash.
  Training runs for 3 epochs with a cosine learning rate schedule and
   a warmup ratio of 0.05.
  The effective batch size is 32 (per-device batch size of 1 with
  gradient accumulation over 8 steps across 4~H20 GPUs).

  \paragraph{RL Stage.}
  We train the SFT-initialized model with DAPO (without KL loss)
  using the veRL framework \citep{sheng2025hybridflow}.
  The batch size is 32 prompts and we generate 8 rollouts per
   prompt, with a train mini-batch size of 16.
  Training runs for 120 steps with a cosine learning rate schedule
  and linear warmup over the first 10 steps.
  The clip ratios are set to $\epsilon_{\text{low}} = 0.2$ and
  $\epsilon_{\text{high}} = 0.28$, and the rollout temperature is
  1.0.

  \paragraph{Compute Resources.}
  The SFT warm-up stage is conducted on 4 NVIDIA H20 GPUs.
  The RL fine-tuning stage is conducted on 8 NVIDIA H20 GPUs, with
  each complete RL training run finishing in approximately 16 hours.

  \begin{table}[h]
  \centering
  \begin{minipage}[t]{0.48\linewidth}
  \centering
  \caption{Hyperparameters for the SFT warm-up stage.}
  \label{tab:sft_hyperparams}
  \begin{tabular}{lc}
  \toprule
  \textbf{Hyperparameter} & \textbf{Value} \\
  \midrule
  Framework            & LLaMA-Factory \\
  Parallelism          & DeepSpeed ZeRO-3 \\
  Number of GPUs       & 4 $\times$ H20 \\
  Training epochs      & 3 \\
  Per-device batch size & 1 \\
  Gradient accum.\ steps & 8 \\
  Effective batch size & 32 \\                                       
  Learning rate        & $1 \times 10^{-5}$ \\
  LR scheduler         & Cosine \\                                   
  Warmup ratio         & 0.05 \\         
  Max sequence length  & 8{,}192 \\
  Max training samples & 5{,}000 \\
  Precision            & BF16 \\
  \bottomrule
  \end{tabular}
  \end{minipage}
  \hfill
  \begin{minipage}[t]{0.48\linewidth}
  \centering
  \caption{Hyperparameters for the RL fine-tuning stage.}
  \label{tab:rl_hyperparams}
  \begin{tabular}{lc}
  \toprule
  \textbf{Hyperparameter} & \textbf{Value} \\
  \midrule
  Algorithm            & DAPO (no KL) \\
  Framework            & veRL \\
  Number of GPUs       & 8 $\times$ H20 \\
  Total training steps & 120 \\
  Prompts per batch    & 32 \\
  Rollouts per prompt  & 8 \\
  Effective batch size & 256 \\
  Mini-batch size      & 16 \\
  Learning rate        & $1 \times 10^{-6}$ \\
  LR scheduler         & Cosine \\
  Warmup steps         & 10 \\
  Clip ratio $\epsilon_{\text{low}}$  & 0.2 \\
  Clip ratio $\epsilon_{\text{high}}$ & 0.28 \\
  Rollout temperature  & 1.0 \\
  \bottomrule
  \end{tabular}
  \end{minipage}
  \end{table}
  
\subsection{Prompts}

\begin{figure}[h]
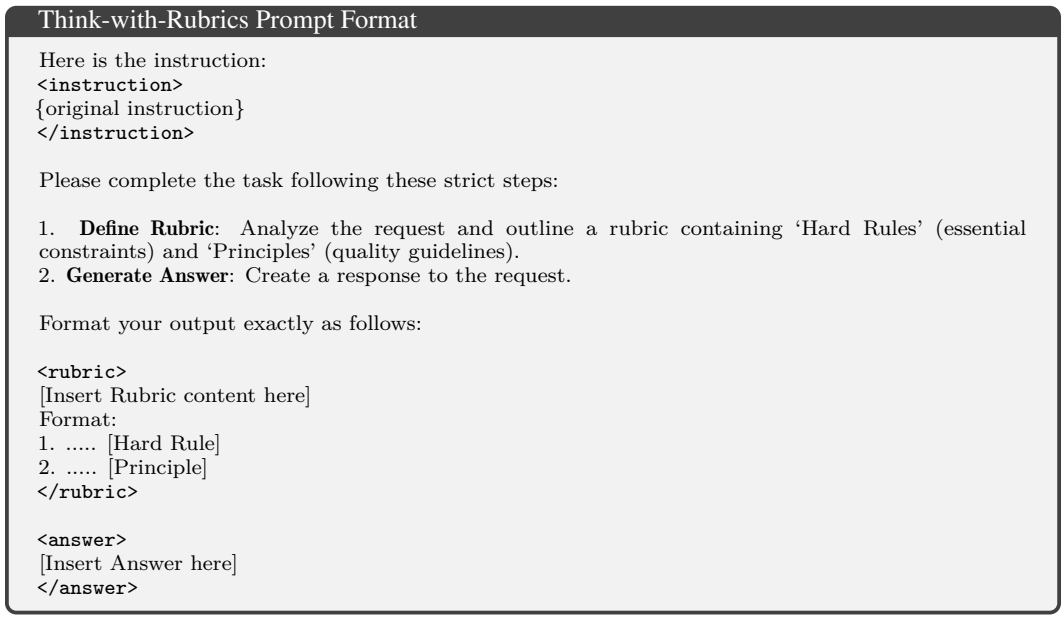

      \centering
      \begin{prompt}{Think-with-Rubrics Prompt Format}

  Here is the instruction:\\
  \texttt{<instruction>}\\
  \{original instruction\}\\
  \texttt{</instruction>}\\

  Please complete the task following these strict steps:\\

  1. \textbf{Define Rubric}: Analyze the request and outline a rubric
   containing
  `Hard Rules' (essential constraints) and `Principles' (quality
  guidelines).\\
  2. \textbf{Generate Answer}: Create a response to the request.\\

  Format your output exactly as follows:\\

  \texttt{<rubric>}\\
  {[Insert Rubric content here]}\\
  Format:\\
  \quad 1. ..... [Hard Rule]\\
  \quad 2. ..... [Principle]\\
  \texttt{</rubric>}\\

  \texttt{<answer>}\\
  {[Insert Answer here]}\\
  \texttt{</answer>}

      \end{prompt}
      \caption{The prompt format used for \textsc{Think-with-Rubrics}
   models. The
      model is explicitly required to first enumerate hard rules and
  quality
      principles as a structured rubric before generating the final
  answer,
      mirroring the \texttt{<rubric>} $\rightarrow$ \texttt{<answer>}
   generation
      paradigm enforced during training.}
      \label{fig:prompt-wrapper}
  \end{figure}

 During inference, we employ two distinct prompt formats
  corresponding to the two                                           
  paradigms evaluated in main result table. For the
  Rubric-as-Reward
  baseline with CoT reasoning (With think setting),
  we append a                                                        
  concise reasoning directive to the original instruction, asking the
   model to first                                                    
  think step by step and then enclose its final answer within
  \texttt{<answer>} tags.
  This format allows the model to perform free-form reasoning prior
  to answering, while
  remaining consistent with the format used during its SFT warm-up
  stage.
  For \textsc{Think-with-Rubrics}, the prompt explicitly instructs
  the model to first
  construct a structured rubric comprising hard rules and quality
  principles, and then
  generate a response conditioned on that rubric, mirroring the
  two-stage generation
  paradigm enforced during training. Both formats are presented in
  Figures~\ref{fig:prompt-think} and \ref{fig:prompt-wrapper}
  respectively.

\subsection{Broader Impacts}
                       
  This work proposes \textsc{Think-with-Rubrics}, a training paradigm
   that improves
  the instruction-following capability and alignment of large
  language models.
  On the positive side, more reliably instruction-following models
  can benefit a wide
  range of downstream applications, including assistive writing
  tools, educational
  systems, and professional productivity software, by producing
  outputs that more
  faithfully respect user intent and constraints. Furthermore, the
  self-evolution
  potential demonstrated in this work suggests a path toward reducing
   dependence on
  costly human annotation, which could democratize the development of
   aligned AI
  systems for resource-constrained settings.

  On the negative side, as with most work aimed at improving general
  instruction
  following, there is a possibility that enhanced compliance with
  user-specified
  constraints could be misused to generate content that more
  precisely satisfies
  undesirable requirements. We note, however, that this risk is not
  specific to our
  approach and is shared by the broader line of work on LLM
  alignment.

 \begin{figure}[h]
      \centering
      \begin{prompt}{Rubric-as-Reward (With Think) Prompt Format}

  \textbf{Instruction: } \{original instruction\}\\

  Please first think step by step and finally put your final answer
  (don't add any
  other output) between \texttt{<answer>} and \texttt{</answer>}.

      \end{prompt}
      \caption{The prompt format used for the Rubric-as-Reward
  baseline with
      CoT reasoning. The model is free to generate an
  unstructured
      reasoning trace before producing the final answer enclosed in
      \texttt{<answer>} tags.}
      \label{fig:prompt-think}
  \end{figure}
\newpage
  
\end{document}